%% file: main.tex
\documentclass[10pt,twocolumn,letterpaper]{article}

\usepackage[pagenumbers]{styles/wacv}

\input{preamble}

\input{sections/0X_FiguresTables}

\definecolor{wacvblue}{rgb}{0.21,0.49,0.74}
\usepackage[pagebackref,breaklinks,colorlinks,allcolors=wacvblue]{hyperref}

\def\wacvPaperID{}
\def\confName{WACV}
\def\confYear{2027}

\title{BiDeMem: Bidirectional Degradation Memory for Explainable Image Restoration}

\author{
    Xinrui Wu\textsuperscript{\rm 1*}, Lichen Huang\textsuperscript{\rm 1*}\\
    \textsuperscript{\rm 1}University of Electronic Science and Technology of China\\
    \textsuperscript{*}Equal contribution.\\
    \textbf{Email:} xinruiwu.wxr@gmail.com, xhghlc@gmail.com
}

\begin{document}
\maketitle
\input{sections/00_Abstract}
\FigMethod
\FigSamples
\input{sections/01_Introduction}
\input{sections/02_Related_Work}
\input{sections/03_Method}
\FloatBarrier
\input{sections/04_Experiments}
\FloatBarrier
\input{sections/05_Discussion}
\input{sections/06_Conclusion}
\FloatBarrier
{
    \small
    \bibliographystyle{styles/ieeenat_fullname_unsrt}
    \bibliography{references/main}
}

\end{document}

%% file: preamble.tex
\usepackage{mathtools}
\usepackage{makecell}
\usepackage{multirow}
\usepackage{array}
\usepackage{placeins}


\newcommand{\ours}{BiDeMem\xspace}
\newcommand{\rankmem}{Rank Memory\xspace}
\newcommand{\bimem}{BiRank Memory\xspace}
\newcommand{\nafnet}{NAFNet\xspace}
\newcommand{\psnr}{PSNR\xspace}
\newcommand{\ssim}{SSIM\xspace}
\newcommand{\yhat}{\hat{y}}
\newcommand{\xhat}{\hat{x}}
\newcommand{\TopK}{\operatorname{TopK}}
\newcommand{\softmax}{\operatorname{softmax}}
\definecolor{wxrcommentpurple}{RGB}{126, 63, 181}



%% file: sections/0X_FiguresTables.tex
\newcommand{\figref}[1]{Figure~\ref{#1}}
\newcommand{\tabref}[1]{Table~\ref{#1}}

\graphicspath{{Figures/}{Figures/main/}{Figures/generated/}{Figures/appendix/}}

\newcommand{\FigMethod}{%
\begin{figure*}[!t]
    \centering
    \includegraphics[width=0.98\textwidth]{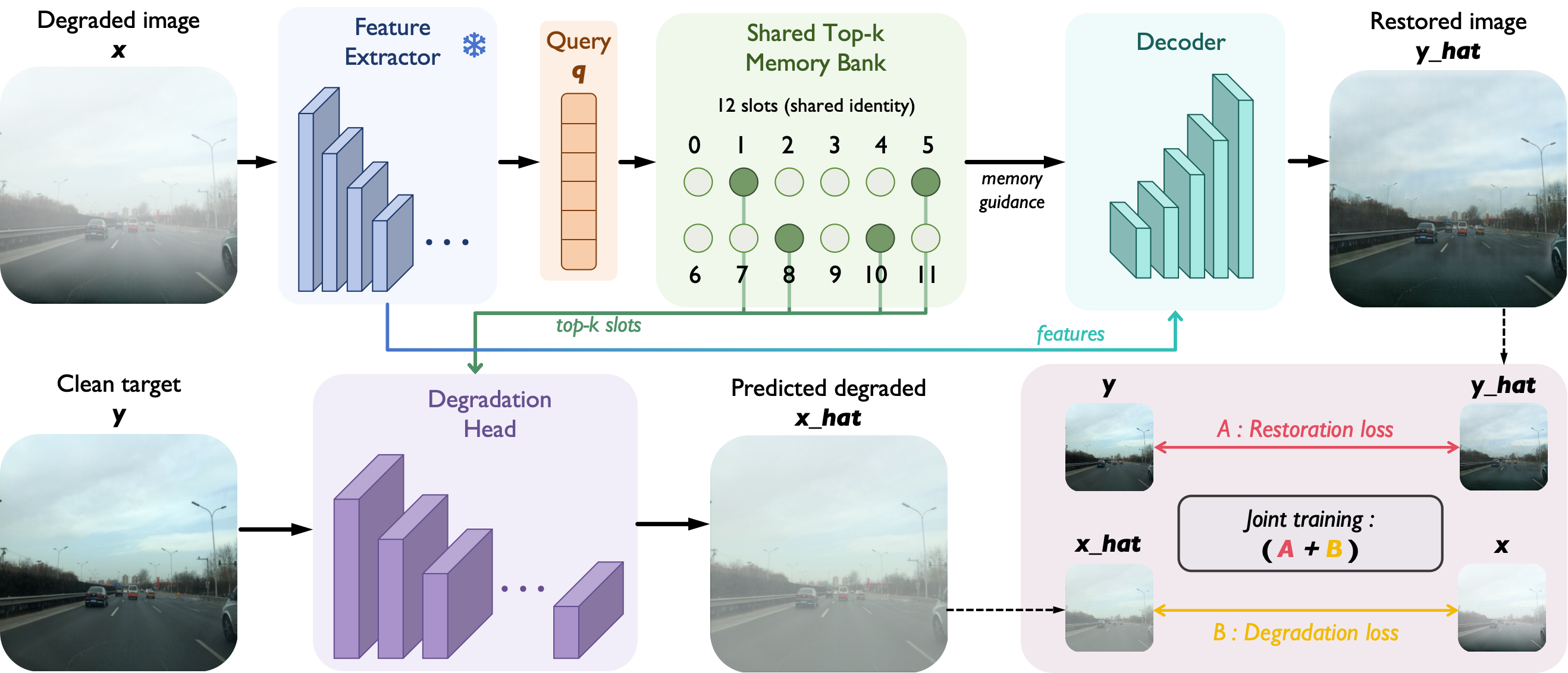}
    \caption{\textbf{Overview of \ours.}
    A degraded image is encoded by a restoration backbone. Evidence from the bottleneck feature and input statistics forms a query that retrieves a compact top-$k$ subset from a degradation memory. The same selected slot identity conditions the restoration decoder during inference and, during training only, drives a forward degradation explanation branch. The figure emphasizes the paper's central claim: a degradation prior should be useful for restoration and falsifiable through reverse degradation and counterfactual interventions.}
    \label{fig:method}
\end{figure*}
}

\newcommand{\FigExplainability}{%
\begin{figure}[!htbp]
    \centering
    \includegraphics[width=\linewidth]{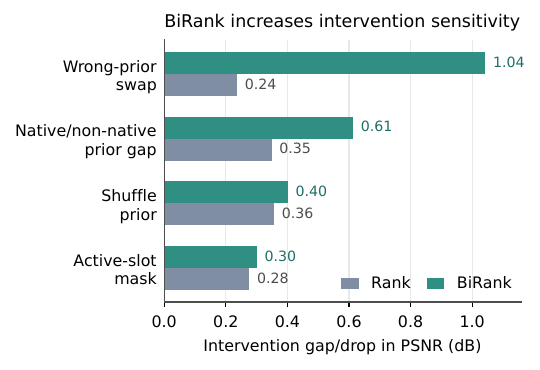}
    \caption{\textbf{Intervention evidence.}
    The bars summarize counterfactual prior tests. Larger gaps or drops mean that the output depends more strongly on the correct retrieved prior. \bimem especially increases wrong-prior sensitivity and the native/non-native prior gap while preserving restoration accuracy.}
    \label{fig:explainability}
\end{figure}
}

\newcommand{\FigSamples}{%
\begin{figure*}[!t]
    \centering
    \includegraphics[width=0.98\textwidth]{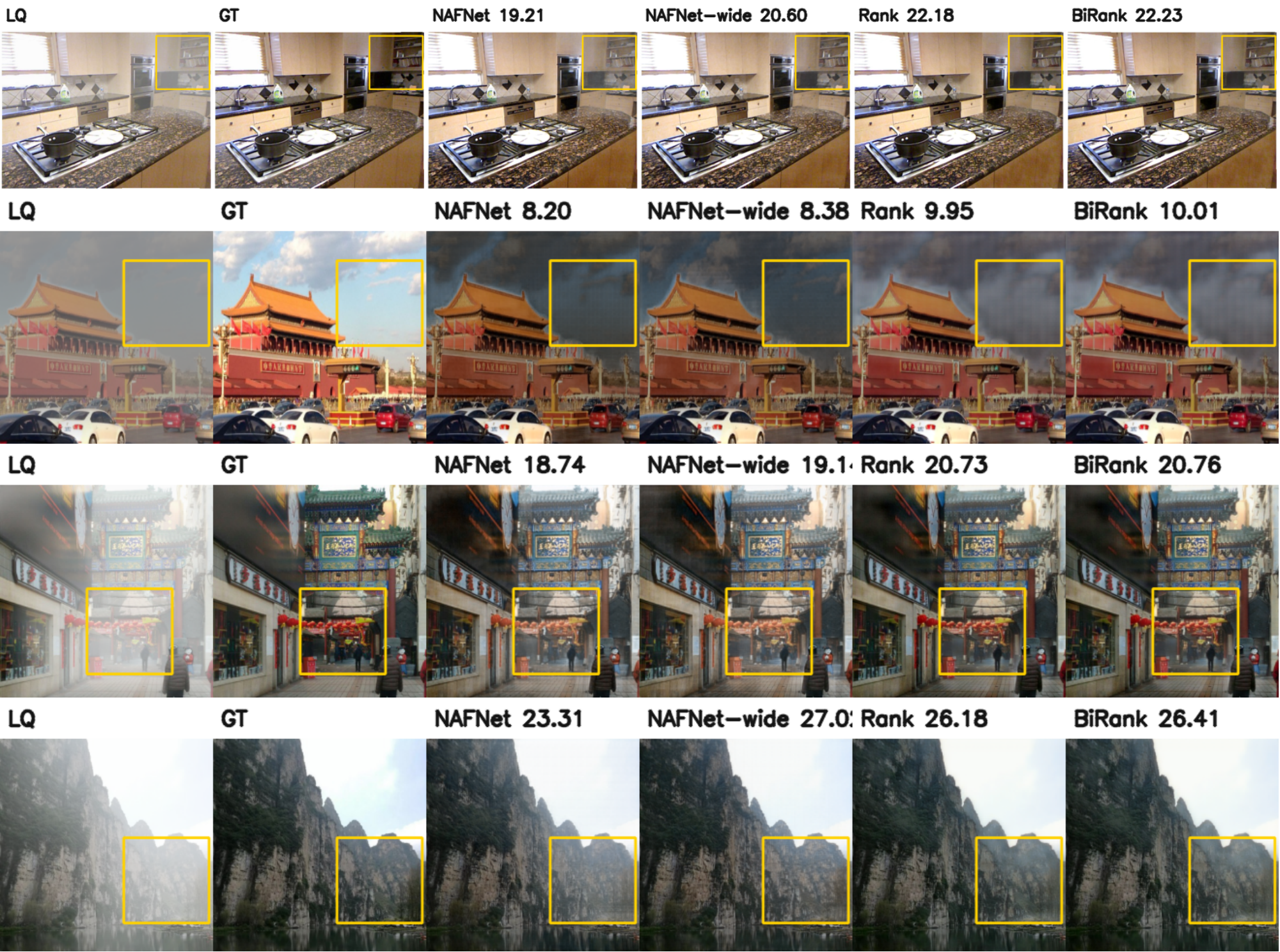}
    \caption{\textbf{Qualitative restoration examples.}
    Each row compares the low-quality input, ground truth, \nafnet, \nafnet-wide, \rankmem, and \bimem. Yellow boxes highlight local regions where the memory variants better recover contrast and structure. The numbers in the headers are per-example \psnr.}
    \label{fig:samples}
\end{figure*}
}

%% file: sections/00_Abstract.tex
\begin{abstract}
Degradation-aware prompts, conditions, and latent priors are increasingly used in image restoration, yet they are usually judged by a single endpoint: whether the restored image obtains higher \psnr. This is a weak test of semantics. A condition can help by adding capacity, acting as a global correction bias, or exploiting dataset shortcuts, without becoming an interpretable degradation prior. We propose \ours, a bidirectional degradation memory for explainable image restoration. A query built from restoration features and input statistics retrieves a compact top-$k$ subset of memory slots. The same selected slot identity supports the restoration path at inference time and a training-only forward-degradation explanation path. The study centers on verifiability in a controlled multi-degradation \nafnet setting. New controls separate the gain from a correction head alone, a dense query prior, and a static global prior: these variants are 0.2588, 0.2586, and 0.2839 dB below BiRank, respectively. Strong residual supervision and a wider degradation head also remain below the full bidirectional memory model. Intervention probes show that BiRank preserves restoration quality while increasing wrong-prior and native-prior sensitivity, framing degradation memory as both a restoration module and a falsifiable explanation mechanism.
\end{abstract}

%% file: sections/01_Introduction.tex
\section{Introduction}
\label{sec:intro}

Image restoration aims to recover a clean image from observations degraded by noise, haze, rain, compression, blur, or their mixtures. It is a central low-level vision problem because restoration quality directly affects both human perception and downstream visual systems. Modern restoration networks have achieved strong progress by improving backbone architectures, enlarging receptive fields, and learning restoration operators that generalize across a range of corruptions~\cite{zhang2017beyond,li2018benchmarking,yang2017deep,zamir2021multi,zamir2022restormer,chen2022simple}. Recent blind and multi-degradation systems further introduce degradation-aware conditions, including prompts, latent representations, contrastive codes, and memory states, so that one model can adapt its restoration behavior to different input degradations~\cite{li2022all,valanarasu2022transweather,potlapalli2023promptir}.

These degradation-aware conditions are useful, but their semantic role is often evaluated indirectly. In many restoration papers, a condition is treated as meaningful when the final restored image obtains higher \psnr or \ssim. Such endpoint evaluation leaves several alternatives unresolved. A condition may encode degradation evidence, yet it may also act as extra model capacity, a global correction bias, a dense prompt vector, or a dataset-specific shortcut. For an explainable degradation prior, a stronger evaluation should ask whether the prior is input-aligned, whether a wrong prior damages restoration, whether only the selected memory slots matter, and whether the same route can support a training signal in the reverse direction from clean target to degraded observation.

We propose \ours, a bidirectional degradation memory for explainable image restoration. The method builds an evidence-guided query from restoration bottleneck features and simple input statistics, retrieves a compact top-$k$ subset of memory slots, and uses the selected slot identity to condition restoration. During inference, the selected slots modulate the restoration decoder and a slot-routed correction head. During training, the same selected slot identity also conditions a forward-degradation consistency branch that reconstructs the degraded observation from the clean target. This bidirectional design turns the retrieved slot identity into an auditable object: the active slots can be masked, replaced, shuffled, compared against non-native priors, and tested under uniform or zero-memory interventions.

We evaluate \ours in a controlled multi-degradation setting built on \nafnet. The goal is to test whether compact degradation memory is useful and verifiable under clear controls. \bimem obtains 29.7529 dB / 0.8865 average \psnr/\ssim over eight seen benchmarks, while correction-head-only, dense-prior, and static/global-prior controls are 0.2588, 0.2586, and 0.2839 dB lower, respectively. More importantly, bidirectional training increases intervention sensitivity: the wrong-prior drop rises from 0.2365 dB for \rankmem to 1.0430 dB for \bimem, and the native/non-native prior gap rises from 0.3484 dB to 0.6134 dB. Additional unseen/mixed tests, low-data adaptation, efficiency measurements, and external-backbone checks define the useful scope and current limitations of the method.

Overall, the contributions of this paper are as follows:
\begin{itemize}[nosep,leftmargin=*]
    \item We formulate degradation memory as a verifiable prior whose value should be assessed by restoration utility, intervention sensitivity, and route consistency.
    \item We propose a compact top-$k$ bidirectional memory whose selected slot identity is shared by restoration and a training-only forward-degradation consistency branch.
    \item We introduce a control suite that separates memory behavior from extra backbone capacity, residual correction heads, dense prompt-like priors, static global priors, degradation supervision, and head capacity.
    \item We show in a controlled \nafnet study that bidirectional memory preserves restoration quality while making the retrieved prior more sensitive to wrong-prior, native-prior, and active-slot interventions, with supportive but setting-dependent external backbone checks.
\end{itemize}

%% file: sections/02_Related_Work.tex
\section{Related Work}
\label{sec:related}

\textbf{Task-specific restoration models.}
Deep image restoration first made rapid progress in task-specific settings. For Gaussian denoising, \emph{DnCNN}~\cite{zhang2017beyond} learned to predict residual noise instead of the clean image, making residual learning a standard design. \emph{FFDNet}~\cite{zhang2018ffdnet} further introduced a noise-level map, which allowed one denoiser to handle multiple noise strengths. \emph{CBDNet}~\cite{guo2019toward} moved toward real photographs by modeling signal-dependent noise. Self-supervised denoisers such as \emph{Noise2Noise}~\cite{lehtinen2018noise2noise} and \emph{Noise2Void}~\cite{krull2019noise2void} relaxed the need for clean targets, but they mainly addressed supervision rather than degradation-prior verifiability.

Similar task-specific progress appears in dehazing, deraining, deblurring, and super-resolution. The \emph{Dark Channel Prior}~\cite{he2010single} estimated haze from hand-crafted image statistics, while \emph{DehazeNet}~\cite{cai2016dehazenet} learned transmission maps end to end. \emph{AOD-Net}~\cite{li2017aod} folded the atmospheric scattering model into a compact restoration network, and \emph{RESIDE}~\cite{li2018benchmarking} provided a widely used dehazing benchmark. In deraining, \emph{JORDER}~\cite{yang2017deep} jointly detected and removed rain streaks, \emph{DID-MDN}~\cite{zhang2018density} used density-aware multi-stream prediction, \emph{DDN}~\cite{fu2017removing} separated image detail from rain structure, \emph{RESCAN}~\cite{li2018recurrent} aggregated recurrent context, and \emph{SPANet}~\cite{wang2019spatial} used spatial attention for real rain. \emph{DeblurGAN}~\cite{kupyn2018deblurgan} and \emph{DeblurGAN-v2}~\cite{kupyn2019deblurgan} introduced adversarial deblurring pipelines, while \emph{EDSR}~\cite{lim2017enhanced} and \emph{RCAN}~\cite{zhang2018image} improved super-resolution through residual scaling and channel attention. These models define strong restoration operators, but their priors are usually tied to a specific degradation or evaluated mainly through endpoint quality.

\textbf{General restoration backbones.}
Recent work has shifted from task-specific modules to general restoration backbones. \emph{MPRNet}~\cite{zamir2021multi} uses multi-stage progressive refinement, allowing earlier stages to guide later restoration. \emph{MIRNet}~\cite{zamir2020learning} learns enriched multi-scale features for real image restoration and enhancement. Transformer-based designs then introduced stronger long-range modeling. \emph{SwinIR}~\cite{liang2021swinir} adapts shifted-window attention to restoration, \emph{Uformer}~\cite{wang2022uformer} builds a U-shaped Transformer, and \emph{Restormer}~\cite{zamir2022restormer} improves high-resolution restoration with efficient attention. \emph{NAFNet}~\cite{chen2022simple} shows that a simplified convolutional backbone can be highly competitive without complex nonlinear activations. \ours uses \nafnet as a controlled backbone because it is strong, simple, and does not already contain a specialized degradation-memory mechanism.

\textbf{Blind and multi-degradation restoration.}
Blind restoration methods ask one model to handle unknown or mixed corruptions. \emph{AirNet}~\cite{li2022all} learns a degradation representation and uses it to guide an all-in-one restoration network. \emph{TransWeather}~\cite{valanarasu2022transweather} targets adverse-weather restoration with a Transformer-based shared model. \emph{PromptIR}~\cite{potlapalli2023promptir} introduces learnable prompts that adapt the restoration network to different degradations. Low-light systems such as \emph{SID}~\cite{chen2018learning} and \emph{Zero-DCE}~\cite{guo2020zero} show another form of degradation-aware enhancement, where the restoration behavior is shaped by image-specific exposure or curve parameters. These methods demonstrate that degradation conditioning can improve a shared model. \ours focuses on the next question: whether the selected condition can be treated as an input-aligned and intervention-sensitive prior.

\textbf{Prompts, modulation, and memory mechanisms.}
Conditioning modules provide the architectural tools for injecting task information into a network. \emph{FiLM}~\cite{perez2018film} modulates features by learned scale and shift parameters, and language-conditioned modulation~\cite{de2017modulating} shows how external evidence can change visual computation. \emph{Dynamic Filter Networks}~\cite{jia2016dynamic} generate filters conditioned on the input, while \emph{PromptIR}~\cite{potlapalli2023promptir} uses prompt tokens as restoration-specific conditions. These designs motivate the FiLM-style conditioning in \ours, but they usually produce dense conditions rather than a small active route.

Memory models make the route more explicit. \emph{Memory Networks}~\cite{weston2014memory} and \emph{End-to-End Memory Networks}~\cite{sukhbaatar2015end} store information in addressable slots. \emph{Neural Turing Machines}~\cite{graves2014neural} and the \emph{Differentiable Neural Computer}~\cite{graves2016hybrid} extend this idea to differentiable external memory. \emph{Slot Attention}~\cite{locatello2020object} learns object-centric slots, and \emph{Product-Key Memory}~\cite{lample2019large} scales retrieval to large memory tables. \ours adapts this slot view to image degradation priors: the memory is compact, top-$k$ retrieval exposes an active subset, and the same selected slot identity is reused across restoration and forward-degradation consistency.

\textbf{Explainability and counterfactual evaluation.}
Explainability methods test whether a prediction depends on meaningful evidence. \emph{Grad-CAM}~\cite{selvaraju2020grad} localizes discriminative image regions, \emph{Integrated Gradients}~\cite{sundararajan2017axiomatic} attributes predictions to input features, and \emph{TCAV}~\cite{kim2018interpretability} measures sensitivity to human-defined concepts. \emph{LIME}~\cite{ribeiro2016should} fits local surrogate explanations, while \emph{SHAP}~\cite{lundberg2017unified} assigns feature contributions through a game-theoretic formulation. Counterfactual explanations~\cite{wachter2017counterfactual} and contrastive explanations~\cite{dhurandhar2018explanations} ask how predictions change when evidence is modified or removed. In restoration, such tests are less commonly applied to the degradation prior itself. \ours therefore evaluates the memory route through prior shuffling, wrong-prior replacement, native/non-native comparison, uniform attention, zero memory, and active/inactive slot masking.

%% file: sections/03_Method.tex
\section{Method}
\label{sec:method}

\subsection{Problem Setup}

Given a degraded image $x\in\mathbb{R}^{3\times H\times W}$ and clean target $y$, a restoration model predicts $\yhat=f(x)$. Standard degradation-conditioned restoration asks whether a condition improves $\yhat$. \ours asks a stronger question: can the same condition be made useful for restoration, identifiable under interventions, and explanatory of the forward degradation process? The deployed path remains a normal restoration path from $x$ to $\yhat$. The bidirectional path is used only during training.

\subsection{Verification Logic}

The method is designed to be tested by controls. We use \nafnet-wide to control parameter count, \emph{\nafnet + correction head} to control the residual correction module, \emph{dense query-FiLM} to control dense prompt-like conditioning without slots, and \emph{static/global prior} to remove input-conditioned retrieval. These controls target the main alternative explanations for a memory-related gain: extra backbone capacity, an additional residual head, dense conditioning, and an input-independent global prior.

\begin{center}
\small
\captionof{table}{\textbf{Baseline map.} Each control blocks one alternative explanation for why memory-like conditioning may help.}
\label{tab:baseline_map}
\begin{tabular}{@{}p{0.54\linewidth}p{0.38\linewidth}@{}}
\toprule
Question & Control \\
\midrule
Is this just more backbone capacity? & \nafnet-wide \\
Is this just a residual head? & \nafnet + correction head \\
Is a dense condition enough? & Dense query-FiLM prior \\
Is top-$k$ retrieval necessary? & Static/global prior \\
Does auxiliary degradation supervision alone close the gap? & Strong residual supervision \\
\bottomrule
\end{tabular}
\end{center}

\subsection{Evidence-Guided Query}

Let $E$ denote the \nafnet feature extractor. From the degraded input we obtain a bottleneck feature $z=E_{\mathrm{bn}}(x)$ and cached skip features. The memory query combines two forms of evidence. First, $z$ is summarized by spatial and frequency-band descriptors, including low-, mid-, and high-frequency FFT statistics. Second, the input image contributes a 22-dimensional observation vector containing RGB statistics, luminance statistics, gradient magnitudes, FFT band energies, and high/low-frequency ratios. The two sources are mapped to a shared 128-dimensional query:
\begin{equation}
    q = \mathrm{norm}\left(g_z(z) + \lambda g_s(s(x))\right), \qquad \lambda=0.25 .
    \label{eq:query}
\end{equation}
The query is deliberately tied to observable degradation evidence; it is not a free image identifier.

\subsection{Compact Top-k Memory}

The memory has $N=12$ slots with dimension $d=128$. A key table $K\in\mathbb{R}^{N\times d}$ is used for addressing, and value tables provide conditioning vectors for the restoration and degradation-explanation heads. Slot scores are normalized dot products with temperature $\tau=0.2$:
\begin{equation}
    r_i = \frac{\langle q, K_i\rangle}{\tau}, \qquad
    \mathcal{S} = \TopK(r, k=4).
\end{equation}
Only the selected slots are active:
\begin{equation}
    a_i =
    \begin{cases}
    \softmax(r_{\mathcal{S}})_i, & i\in\mathcal{S},\\
    0, & \text{otherwise}.
    \end{cases}
    \label{eq:attention}
\end{equation}
The active set $\mathcal{S}$ is the object we test. A useful memory should not distribute its effect uniformly over all slots, nor should it collapse into a static global vector.

\subsection{Restoration Path}

The restoration value read from the active slots produces a prior $p_r$. This prior modulates the \nafnet decoder by FiLM at all decoder stages, while the decoder still receives the bottleneck and skip features from the image. Thus memory conditions the restoration process but does not replace image evidence. The decoder output is a base restoration $b$.

We then use a slot-routed correction head. It receives $\mathrm{concat}(x,b,x-b)$ and predicts a global correction $c_g$ and slot-indexed basis corrections $B_i$. The final image is
\begin{equation}
    \yhat = b + c_g + \sum_{i=1}^{N} w_i B_i ,
    \label{eq:restore}
\end{equation}
where $w=\softmax(\log(a+\epsilon)/0.5)$. This makes the output explicitly depend on the retrieved active slots.

\subsection{Bidirectional Explanation Path}

\bimem adds a training-only path that uses the same selected slot identity $\mathcal{S}$ to explain degradation. A degradation value read from the selected slots gives $p_d$. A forward degradation head receives the clean target $y$ and $p_d$ and predicts
\begin{equation}
    \xhat = y + h_d(y,p_d).
\end{equation}
An explanation correction head and consistency gate refine this prediction. The branch is supervised so that $\xhat$ matches $x$ in pixel, frequency, low-frequency, and statistic spaces. The important constraint is identity sharing: the slots used to help restoration must also support a plausible forward degradation explanation.

\subsection{Objectives}

All variants use a restoration loss
\begin{equation}
    \mathcal{L}_{\mathrm{pix}}=\|\yhat-y\|_1 .
\end{equation}
We also use a prior ranking loss comparing the native prior to wrong priors from a cross-degradation queue:
\begin{equation}
    \mathcal{L}_{\mathrm{rank}} =
    \max(0, e_{\mathrm{native}} + m - e_{\mathrm{wrong}}),
    \label{eq:rank}
\end{equation}
with margin $m=0.001$. \rankmem uses the restoration path and ranking loss. \bimem adds degradation and explanation terms:
\begin{equation}
    \mathcal{L}_{\mathrm{BiRank}}
    = \mathcal{L}_{\mathrm{pix}}
    + \alpha\mathcal{L}_{\mathrm{rank}}
    + \beta\mathcal{L}_{\mathrm{deg}}
    + \gamma\mathcal{L}_{\mathrm{explain}} .
    \label{eq:birank}
\end{equation}

%% file: sections/04_Experiments.tex
\section{Experiments}
\label{sec:experiments}

\subsection{Protocol}

We use a controlled multi-degradation setting with denoising, deraining, and dehazing. Denoising uses BSD400~\cite{martin2001bsd} and WED images with Gaussian noise levels $\sigma\in\{15,25,50\}$; deraining uses 5,200 paired samples; dehazing uses 5,500 synthetic/clean pairs following common dehazing benchmarks~\cite{li2018benchmarking}. Testing uses BSD68~\cite{martin2001bsd}, Urban100~\cite{huang2015urban100}, Rain100L~\cite{yang2017deep}, and Haze4K. Unless stated otherwise, training uses 30k iterations, AdamW, learning rate $10^{-4}$ cosine-decayed to $10^{-7}$, weight decay $10^{-3}$, $192\times192$ patches, global batch size 4, and seeds(42/123/2026).

The controlled backbone is \nafnet~\cite{chen2022simple} with width 32 and block layout $[1,1,1,28]$ encoder, one middle block, and $[1,1,1,1]$ decoder. \nafnet-wide uses width 40. Memory variants use 12 slots, top-4 retrieval, 128-dimensional values, and train the query, memory, correction, degradation, explanation, and gate modules while keeping the backbone restoration parameters frozen.

The experimental logic follows \tabref{tab:baseline_map}. We first establish controlled restoration utility, then test supporting generalization, low-data memory reuse, and efficiency, before turning to the mechanism controls and intervention evidence that form the main explanation claim.

\begin{center}
\small
\captionof{table}{\textbf{Controlled restoration context.} Average over eight seen tests. Main rows use the three-seed protocol.}
\label{tab:restoration_context}
\resizebox{\linewidth}{!}{%
\begin{tabular}{lccc}
\toprule
Method & Params & \psnr/\ssim & Role \\
\midrule
\nafnet & 17.11M & 29.4909 / 0.8757 & Backbone \\
\nafnet-wide & 26.65M & 29.5962 / 0.8783 & Capacity \\
\rankmem & 18.02M & 29.7467 / 0.8861 & Slot memory \\
\bimem & 18.02M & \textbf{29.7529} / \textbf{0.8865} & Bidirectional \\
\bottomrule
\end{tabular}}
\end{center}

\subsection{Controlled Restoration Utility}

\tabref{tab:restoration_context} gives the controlled restoration context. This is the entry point rather than the final claim: the memory must be useful enough to matter, but restoration accuracy alone cannot prove that the slots encode a meaningful degradation prior. In this controlled \nafnet setting, \bimem is slightly above \rankmem in average restoration quality while also adding bidirectional explanation.

The aggregate seen-benchmark result should be read as controlled restoration utility rather than as a claim of uniform dominance on every degradation. The purpose of this experiment is to establish that the memory route is useful enough to study; the following controls and interventions then test whether this route behaves like a meaningful degradation prior.

\figref{fig:samples} provides qualitative examples. The memory variants generally restore sharper local contrast and clearer structures than the controlled \nafnet baselines, while the remaining failure modes motivate the more cautious positioning of \ours as an explainable degradation-memory study.

\subsection{Supporting Generalization and Reuse}

\begin{center}
\centering
\small
\captionof{table}{\textbf{Generalization summary.} Average over ten unseen/mixed tests.}
\label{tab:generalization_summary}
\begin{tabular}{lcc}
\toprule
Method & \psnr/\ssim & $\Delta$ \psnr \\
\midrule
\nafnet & 24.7557 / 0.7926 & -- \\
\nafnet-wide & 24.7365 / 0.7926 & -0.0192 \\
\rankmem & 24.9539 / 0.8053 & +0.1982 \\
\bimem & \textbf{24.9556} / \textbf{0.8056} & \textbf{+0.1999} \\
\bottomrule
\end{tabular}
\end{center}

We use unseen and mixed corruptions as supporting evidence that the memory extends beyond seen benchmark labels. Across ten tests including JPEG compression, unseen noise levels, and haze/rain mixtures, \rankmem improves the \nafnet average by +0.1982 dB and \bimem by +0.1999 dB, with wins on seven of ten tests. These gains remain secondary to the intervention evidence, and \tabref{tab:generalization_summary} reports the compact summary.

\begin{center}
\begin{minipage}{\linewidth}
\centering
\scriptsize
\captionof{table}{\textbf{Low-data adaptation.} Average \psnr/\ssim over eight seen benchmarks.}
\label{tab:low_data_adaptation}
\resizebox{\linewidth}{!}{%
\begin{tabular}{lccccc}
\toprule
Data & \nafnet & \rankmem & \bimem & Rank gain & BiRank gain \\
\midrule
10\% & 28.9354 / 0.8669 & 29.7159 / 0.8855 & \textbf{29.7309 / 0.8858} & +0.7805 & +0.7954 \\
25\% & 29.0327 / 0.8696 & 29.7308 / 0.8857 & \textbf{29.7387 / 0.8859} & +0.6981 & +0.7060 \\
50\% & 28.7756 / 0.8640 & 29.7371 / 0.8858 & \textbf{29.7462 / 0.8861} & +0.9615 & +0.9706 \\
\bottomrule
\end{tabular}}
\end{minipage}
\end{center}

With a pretrained compact memory, low-data adaptation at 10\%, 25\%, and 50\% training data improves over same-protocol \nafnet trained from scratch by +0.7805, +0.6981, and +0.9615 dB for \rankmem, and by +0.7954, +0.7060, and +0.9706 dB for \bimem. \tabref{tab:low_data_adaptation} reports the full summary. This comparison measures memory reuse under limited training data.

\subsection{Cost}

\begin{center}
\centering
\small
\captionof{table}{\textbf{Efficiency on $256\times256$ inputs.} \ours is compact in parameters but not faster.}
\label{tab:efficiency}
\begin{tabular}{lccc}
\toprule
Method & Params & FLOPs & Latency \\
\midrule
\nafnet & 17.11M & 31.91G & 26.56 ms \\
\nafnet-wide & 26.65M & 49.62G & 39.46 ms \\
\rankmem & 18.02M & 45.88G & 41.77 ms \\
\bimem & 18.02M & 45.88G & 41.90 ms \\
\bottomrule
\end{tabular}
\end{center}

\tabref{tab:efficiency} reports cost: \ours is compact in parameters but not faster.

\subsection{Memory Design Controls}

\tabref{tab:memory_controls} isolates the memory mechanism under the follow-up protocol. The correction-head-only and dense-prior variants are almost identical, and both are about 0.26 dB below BiRank. Thus, the result is not explained by simply adding the residual correction head or by replacing slots with a dense query-conditioned FiLM prior. Static/global prior is still weaker, showing that input-conditioned retrieval is necessary.

The two strongest negative controls are the correction-only and dense-prior variants. They are nearly tied with each other and far below BiRank, which is exactly the pattern expected if the useful component is not the correction head itself and not a generic dense prompt. The strong-residual and wide-degradation variants are also informative: both recover part of the gap relative to the weakest controls, but remain about 0.19 dB below BiRank. This suggests that degradation supervision or head capacity alone is insufficient without the full bidirectional memory route.

\subsection{Bidirectional Intervention Evidence}

\FigExplainability

\tabref{tab:memory_controls} and \tabref{tab:explainability_controls} answer different questions. \tabref{tab:memory_controls} asks whether the restoration gain can be explained away by capacity, correction heads, dense priors, or static priors. \tabref{tab:explainability_controls} asks whether the retrieved prior behaves like the claimed degradation evidence under interventions. This is the central table for the paper: the small average-\psnr difference is less important than whether the prior fails under counterfactual use. Compared with \rankmem, \bimem raises the wrong-prior drop from 0.2365 dB to 1.0430 dB and the native-prior gap from 0.3484 dB to 0.6134 dB.

The active/inactive masking results support compact reliance. Masking active slots reduces performance, whereas masking inactive slots has essentially no effect. This is the behavior expected from a top-$k$ memory and would be difficult to interpret for an unconstrained dense condition.

\begin{table*}[!tbp]
\centering
\small
\setlength{\tabcolsep}{5pt}
\caption{\textbf{Memory design controls on \nafnet, seed 42.} These rows are mechanism tests, not benchmark comparisons.}
\label{tab:memory_controls}
\begin{tabular}{@{}p{0.22\textwidth}ccp{0.30\textwidth}@{}}
\toprule
Variant & Avg. \psnr/\ssim & $\Delta$ \psnr vs. \bimem & Main conclusion \\
\midrule
\bimem & \textbf{29.7529} / \textbf{0.8865} & -- & Reference \\
Correction head only & 29.4941 / 0.8792 & -0.2588 & Head alone insufficient \\
Dense query prior & 29.4943 / 0.8796 & -0.2586 & Dense prior insufficient \\
Static/global prior & 29.4690 / 0.8738 & -0.2839 & Retrieval matters \\
Strong residual supervision & 29.5637 / 0.8806 & -0.1892 & Supervision alone insufficient \\
Wide degradation head & 29.5583 / 0.8814 & -0.1946 & Capacity is secondary \\
\bottomrule
\end{tabular}

\vspace{0.45em}
\centering
\small
\caption{\textbf{Bidirectional and intervention controls.} Larger gaps/drops indicate stronger dependence on the correct prior.}
\label{tab:explainability_controls}
\resizebox{\textwidth}{!}{%
\begin{tabular}{lcccccc}
\toprule
Method & Avg. \psnr & Shuffle gap & Wrong-prior drop & Native gap & Active-mask drop & Inactive-mask drop \\
\midrule
\rankmem & 29.7467 & 0.3570 & 0.2365 & 0.3484 & 0.2764 & -0.0004 \\
\bimem & \textbf{29.7529} & \textbf{0.4013} & \textbf{1.0430} & \textbf{0.6134} & \textbf{0.3008} & -0.0004 \\
\bottomrule
\end{tabular}}

\vspace{0.45em}
\centering
\small
\caption{\textbf{External backbone reality checks.} Average \psnr/\ssim over the seen benchmark suite. ``Ctrl.'' denotes the available parameter/capacity control. The results provide supportive but setting-dependent evidence beyond the controlled \nafnet backbone.}
\label{tab:external_backbones}
\begin{tabular}{p{0.30\textwidth}cccc}
\toprule
Backbone setting & Base & Ctrl. & \rankmem & \bimem \\
\midrule
AirNet~\cite{li2022all} fine-tuned & 29.8895 / 0.8974 & 30.3699 / 0.8957 & 30.3368 / 0.8975 & \textbf{30.4178} / \textbf{0.8988} \\
AirNet~\cite{li2022all} from scratch & 28.4939 / 0.8681 & 28.6705 / 0.8709 & 28.7314 / 0.8708 & \textbf{28.7459} / \textbf{0.8730} \\
PromptIR~\cite{potlapalli2023promptir} fine-tuned & 29.3825 / 0.9024 & 29.9118 / 0.9005 & 30.1703 / \textbf{0.9039} & \textbf{30.6290} / 0.9032 \\
PromptIR~\cite{potlapalli2023promptir} from scratch & 28.4866 / 0.8564 & 28.7411 / 0.8618 & 28.5491 / 0.8585 & \textbf{28.7767} / \textbf{0.8626} \\
\bottomrule
\end{tabular}
\end{table*}

\subsection{External Backbone Checks}

We also test transfer to stronger degradation-aware backbones. \tabref{tab:external_backbones} reports matched fine-tuned and from-scratch variants for AirNet~\cite{li2022all} and PromptIR~\cite{potlapalli2023promptir}. \bimem gives the best \psnr in all four settings, improving the matched bases by +0.5283/+0.2520 dB on AirNet and +1.2465/+0.2901 dB on PromptIR. Since controls remain competitive and PromptIR fine-tuning gives \rankmem slightly higher \ssim, we treat this as supportive, setting-dependent evidence.

%% file: sections/05_Discussion.tex
\section{Discussion and Limitations}
\label{sec:discussion}

\ours studies degradation memory from the perspective of verification rather than endpoint restoration quality alone. The controlled \nafnet results show that a compact top-$k$ memory can provide a useful restoration prior, while the intervention results show that the retrieved route is measurably sensitive to wrong, shuffled, and non-native priors. This combination is important for explainable restoration: the memory is not only a conditioning vector that improves an output image, but also an addressable route whose behavior can be tested.

The bidirectional design adds an interpretability constraint to this route. By asking the same selected slot identity to support restoration and a training-only forward degradation explanation, \bimem makes the memory pathway more auditable. This constraint is not intended to maximize \psnr in every setting; decoupling restoration and degradation slots may be useful for supplementary analysis or task-specific optimization. The advantage of the shared identity is that counterfactual probes evaluate the same route used by restoration, making the learned prior easier to inspect.

The current design also has clear limitations. \ours is parameter-compact but not compute-efficient: the memory and correction heads add latency relative to the base network. In addition, the compact global slot design may not be the best form for every restoration backbone or degradation mixture. These limitations indicate that the present model should be viewed as a verification-oriented degradation-memory design, not as a universal restoration operator.

The AirNet and PromptIR checks provide encouraging but still setting-dependent external evidence. \bimem improves \psnr over the matched base and capacity-control variants in all four external settings, suggesting that bidirectional memory can transfer beyond \nafnet when jointly adapted. At the same time, the gains vary across backbones and training settings, scratch controls remain competitive, and \ssim is not always maximized by \bimem. Future work should therefore study backbone-aware memory size, insertion points, and training objectives that preserve verifiability while improving efficiency.

%% file: sections/06_Conclusion.tex
\section{Conclusion}
\label{sec:conclusion}

We introduced \ours, a bidirectional degradation memory for explainable image restoration. The method retrieves compact top-$k$ memory slots from observable degradation evidence and uses the same selected slot identity to condition restoration and a training-only forward degradation explanation branch. Across controlled restoration, mechanism controls, intervention probes, low-data adaptation, and external backbone checks, the results support a verification-oriented view of degradation priors: a useful prior should improve restoration, remain aligned with the input degradation, and respond predictably under counterfactual interventions. \ours is not intended as a final all-purpose restoration backbone; rather, it provides a concrete design and evaluation protocol for making degradation memory auditable. We hope this perspective encourages future restoration models to treat learned degradation conditions as testable objects, not only as latent variables that improve endpoint image quality.